\pdfoutput=1

\documentclass[11pt]{article}

\usepackage[final]{acl}

\usepackage{times}
\usepackage{latexsym}

\usepackage{graphicx,xcolor,tikz}
\usepackage{booktabs,multirow}
\usepackage{colortbl}
\usepackage{amsmath}
\usepackage{amssymb}
\usepackage{pifont}
\usepackage{subfigure}
\newcommand{\cmark}{\ding{51}}
\newcommand{\xmark}{\ding{55}}

\usepackage[T1]{fontenc}

\usepackage[utf8]{inputenc}

\usepackage{microtype}

\usepackage{inconsolata}

\title{Detecting Response Generation Not Requiring Factual Judgment}

\author{%
  Ryohei\,Kamei$^{1}$ \quad
  Daiki\,Shiono$^{1}$ \quad
  Reina\,Akama$^{1,2}$ \quad
  Jun\,Suzuki$^{1,2}$ \\
  $^{1}$ Tohoku University \quad
  $^{2}$ RIKEN \\
  \texttt{\{ryohei.kamei.s4, daiki.shiono.s1\}@dc.tohoku.ac.jp}, \\
  \texttt{\{akama, jun.suzuki\}@tohoku.ac.jp} \\
}

\begin{document}
\maketitle
\begin{abstract}
With the remarkable development of large language models (LLMs), ensuring the factuality of output has become a challenge.
However, having all the contents of the response with given knowledge or facts is not necessarily a good thing in dialogues.
This study aimed to achieve both attractiveness and factuality in a dialogue response for which a task was set to predict sentences that do not require factual correctness judgment such as agreeing, or personal opinions/feelings.
We created a dataset, dialogue dataset annotated with fact-check-needed label (DDFC), for this task via crowdsourcing, and classification tasks were performed on several models using this dataset.
The model with the highest classification accuracy could yield about $88$\% accurate classification results.
\end{abstract}

\section{Introduction}
Large language models (LLMs) have undergone considerable development and can solve various natural language processing tasks. However, they output content that is different from the fact, i.e., hallucination, making it difficult to ensure the factuality of the output~\citep{zha-etal-2023-alignscore, dixit-etal-2023-improving, huang-etal-2023-zero}.

Although hallucination in dialogue systems using LLMs has been extensively studied, they focused on methods for detecting/suppressing hallucinations and investigated the causes of their occurrence~\citep{dziri-etal-2022-origin, sun-etal-2023-towards, ji-etal-2023-rho}.
Wizard of Wikipedia (WoW), a knowledge-based dialogue dataset created by \citet{dinan2018wizard}, contains many subjective opinions and feelings of the speaker.
\citet{dziri-etal-2022-faithdial} labeled utterances in WoW datasets that contained subjective opinions and feelings as hallucinations and showed that models fine-tuned on WoW datasets produce more hallucinations.
However, for open-domain dialogue systems such as chatbots, unlike systems in other fields such as summarization or machine translation, not all output in a response are based on a given input or knowledge.
To promote smooth dialogue and increase engagement, expressing personal feelings and opinions is important~\citep{10.1145/3383123}.
Moreover, the tolerance of factual correctness regarding the response of these contents is high~\citep{10.1145/3571730}.

\begin{figure}[t]
\includegraphics[width=\columnwidth]{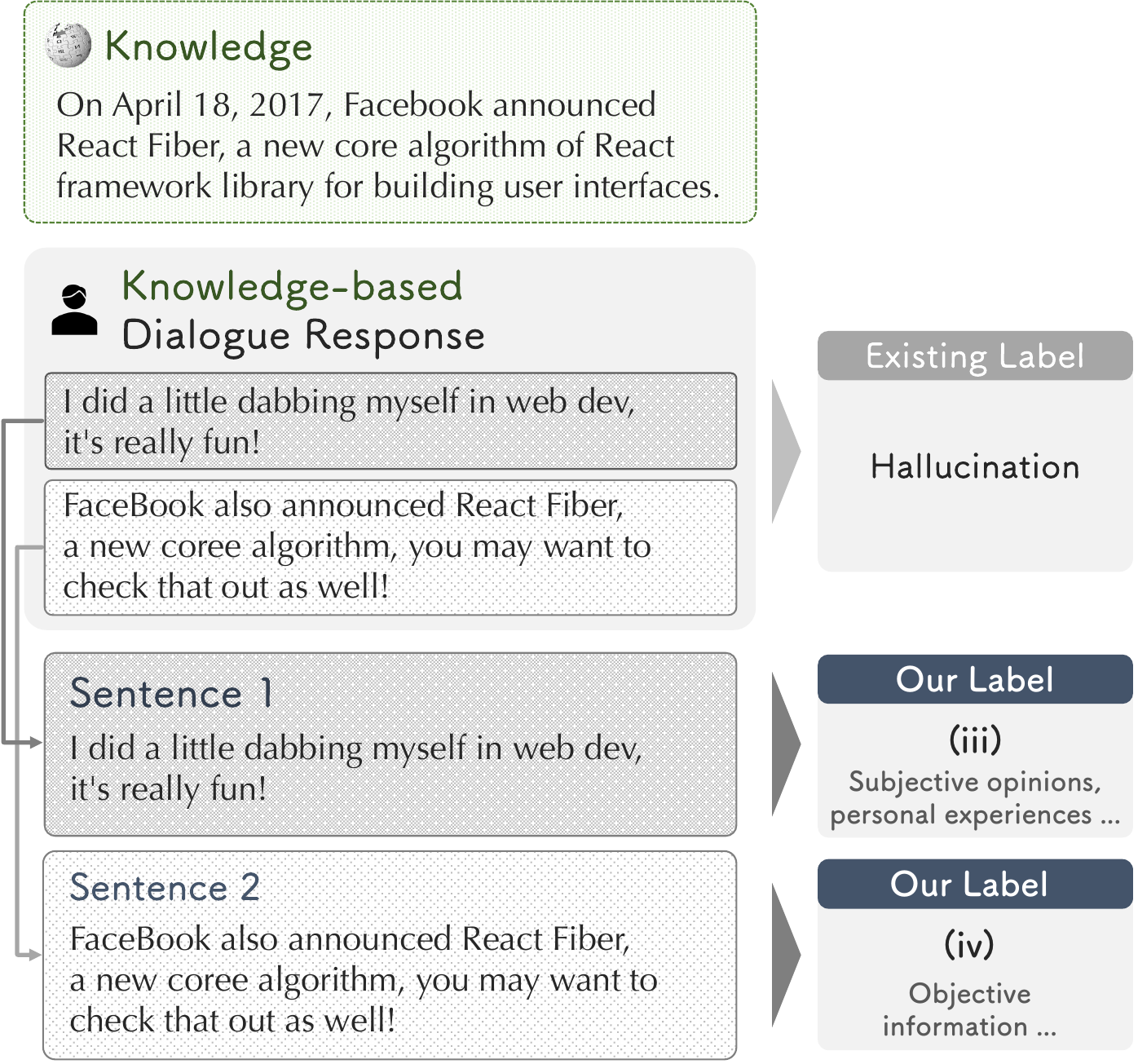}
\caption{Overview of the study and the collected dataset, DDFC. The existing dialogue responses based on knowledge are divided into sentences. Each sentence was annotated labels according to its type and used in a classification task.}
\label{fig:overview}
\end{figure}

\begin{figure*}[t]
    \includegraphics[width=2\columnwidth]{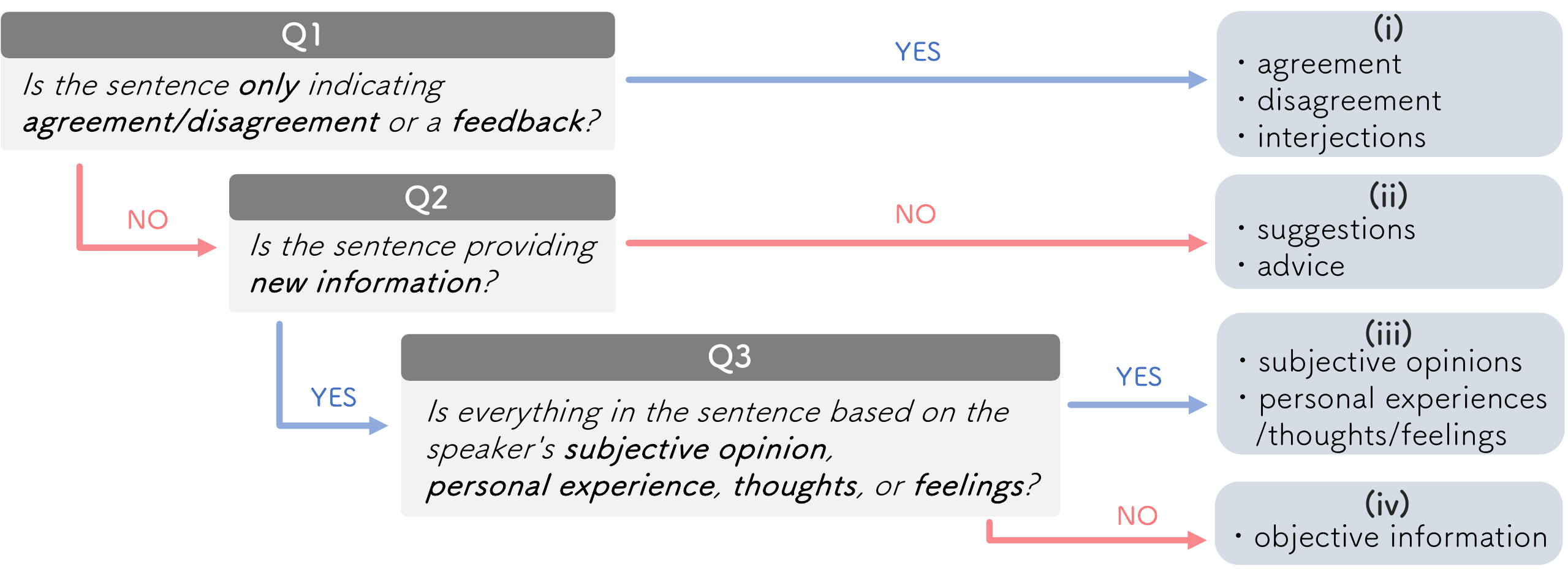}
    \caption{Flowchart of annotation by Amazon Mechanical Turk to construct DDFC.}
    \label{fig:AMT_Flow}
\end{figure*}

To address these issues, we propose that sentences not requiring factual correctness judgment should be detected and removed before judgment (hallucination detection) during response generation in dialogue systems.
By detecting such sentences first and judging the factual correctness of remaining sentences, responses that maintain the attractiveness of the dialogue can be generated while ensuring the factuality of the dialogue.

First, we set the task of detecting sentences that do not require factual correctness judgment, and created a new dataset.
Then, the dataset was validated using classification models.
Figure~\ref{fig:overview} overviews the created dataset, \textbf{dialogue dataset annotated with fact-check-needed label (DDFC)}.
The construction method and contents of DDFC are described in Section~\ref{sec:DDFC_dataset}.

\section{Related Work}

\subsection{Hallucination Detection}

Hallucinations from an LLM output must be detected to improve the reliability of the generated output and apply LLMs to real-world applications.
\citet{guerreiro-etal-2023-optimal} detected hallucinations in machine-translated outputs by formulating them using optimal transport based on the insight that responses containing hallucinations are distant from the source sentences.
Similarly, \citet{dale-etal-2023-detecting} detected hallucinations by evaluating the contribution of the source sentence to the generated sentence.
Various other methods for detecting hallucinations have been proposed in many fields such as summarization and question answering~\cite{choubey-etal-2023-cape, sadat-etal-2023-delucionqa}.

\subsection{Hallucination in Dialogue System}

Detection and suppression of hallucinations are crucial for constructing dialogue systems~\cite{dziri-etal-2022-faithdial}. 
\citet{shuster-etal-2021-retrieval-augmentation} suppressed hallucinations by augmenting a dialogue system with a module that retrieved relevant knowledge. 
\citet{dziri-etal-2021-neural} also proposed a dialogue system that could modify hallucinations in the generated responses by querying the knowledge graph.

\subsection{Knowledge-Grounded Dialogue Dataset}

Knowledge-based dialogue datasets have been created to generate informative and reliable responses by leveraging external knowledge~\cite{xue-etal-2023-improving} such as WoW~\cite{dinan2018wizard}. 
The WoW dataset contains dialogues between an apprentice, an information seeker, and a wizard who responds based on his knowledge of Wikipedia. 
CMU-DOG is another dataset containing conversations based on Wikipedia articles about movies given as knowledge ~\cite{zhou2018dataset}, and TOPICAL-CHAT is a knowledge-based dialogue dataset on eight broad topics~\cite{gopalakrishnan19_interspeech}.

\section{DDFC dataset}
\label{sec:DDFC_dataset}

The DDFC dataset created herein contained external knowledge, responses based on external knowledge, responses split by sentences, sentence labels based on discourse acts, and labels to determine whether factual correctness judgments are required.
We used four types of labels, and crowdworkers assigned them through annotation based on the flowchart (Figure~\ref{fig:AMT_Flow}).

\subsection{Idea}

The FaithDial created by \citet{dziri-etal-2022-faithdial} was based on WoW, wherein a response was labeled as hallucination if it contained information not supported by the given knowledge.
In other words, if the speaker’s subjective opinion, personal experience, thoughts, or feelings are included in the response, it is labeled as hallucination in this dataset.
However, the WoW dataset was created based on this instruction: ``use the given knowledge to provide an appropriate response, rather than simply parrot it, and, if possible, present relevant knowledge in a fun and engaging way''~\cite{dinan2018wizard}. 
Moreover, to evaluate the chatbot system output, not only ``usefulness''  by providing information but also metrics such as ``whether the user wants to talk again,'' ``whether the user is interested'' are used~\cite{inaba_2019}.

Thus, generating utterances based on given knowledge and drawing the users’ interest and empathy by expressing personal opinions and feelings are crucial for dialogue systems. 
Therefore, the knowledge-based dialogue dataset was annotated with a new label that indicated whether a factual correctness judgment was required.

\subsection{Construction of the dataset}

\paragraph{Base dataset of DDFC.}

The dialogue responses based on external knowledge in the FaithDial were labeled after splitting them into sentences. 
FaithDial labels the responses of the Wizard (generates responses based on a given Wikipedia article) with hallucination and dialogue act labels in the WoW dataset.

\paragraph{Sentence split for label annotation.}

In the DDFC, FaithDial responses were split by \{`.', `!', `?', `…'\} to label them in one-sentence units.

\begin{table}[t]
\centering
\small
\begin{tabular}{lrrc}
\toprule
    &  \# of sample & rate(\%) & included \\ 
\midrule
three labels matched &  $815$ & $60.0$ & \cmark \\ 
two labels matched &  $502$ & $36.9$ & \cmark \\
no matched &  $42$  & $3.1$ & \xmark \\
\bottomrule
\end{tabular}
\caption{The label match rate of Crowdworker when annotating DDFC dataset. Since there were only a few instances of no match, the validity of the data collection method was considered high. Sentences with no match were excluded from the dataset.}
\label{tab:annotater_label_match}
\end{table}

\begin{table}[t]
\centering
\small
\begin{tabular}{clrr}
\toprule
 & explanation & \# of sample & rate(\%)\\ \midrule
$(\mathrm{i})$ & agreement, feedback etc. & $141$ & $10.7$\\ 
$(\mathrm{ii})$ & proposal, adivice etc. & $110$ & $8.4$\\
$(\mathrm{iii})$ & subjective opinions etc. & $540$ & $41.0$\\
$(\mathrm{iv})$ & objective info etc. & $526$ & $39.9$\\
\bottomrule
\end{tabular}
\caption{The Number of samples and the percentage of each label in DDFC we created. Sentence label (iii) and (iv) each account for approximately $40$\% of the total.}
\label{tab:dataset_labels}
\end{table}

\begin{table}[t]
    \centering
    \small
    \begin{tabular}{lrr}
    \toprule
    parameter & encoder & decoder \\ 
    \midrule
    number of epochs & $5$ & $2$ \\
    global batch sizes& $64$ & $32$ \\
    optimizer & AdamW & AdamW \\
    learning rate & $5.0\times10^{-4}$ & $5.0\times10^{-5}$ \\
    scheduler & cosine & cosine \\
    max length & $256$ & $1,024$ \\
    \bottomrule
    \end{tabular}
    \caption{Fine-tuning settings for the classification models used in this study.}
    \label{tab:finetuning_settings}
\end{table}

\paragraph{Label types.}

Sentence labels were created with reference to the discourse act tag in the “Corpus of Everyday Japanese Conversation” created by the National Institute for Japanese Language and Linguistics~\cite{9041235}.
We used the following four types of labels: (i) agreement, disagreement, interjections, etc.; (ii) suggestions, advice, etc.; (iii) subjective opinions, personal experiences/thoughts/feelings, etc.; and (iv) objective information, etc.
Responses that are labeled as (i), (ii), and (iii) were considered dialogue acts intended to attract user interest or increase the attractiveness of the dialogue response.
Therefore, they are acceptable even if they are not based on given knowledge and were labeled as not required factual correctness judgment.
In contrast, responses labeled as (iv) that provided objective information must be appropriately based on the given knowledge; therefore, they were assigned the label of requiring a factual correctness judgment.

\begin{table*}[t]
\centering
\small
\begin{tabular}{lcccrrrr}
\toprule
            model & architecture & parameter size & fine-tuning & accuracy  & precision & recall & F1-Score \\ \midrule
$\text{GPT-3.5}$ & decoder & no data & \xmark &   $57.73$  &   $58.17$  &   $96.74$  &   $72.65$  \\
$\text{GPT-4}$ & decoder & no data & \xmark &   $57.73$  &   $58.99$  &   $89.13$  &   $71.00$  \\
$\text{Llama 2}_{\text{Chat 7B}}$ & decoder & $7$B & \xmark &   $58.99$  &   $58.60$  &   $\textbf{100.0}$  &   $73.90$  \\
$\text{Llama 2}_{\text{Chat 7B}}$ & decoder & $7$B & \cmark &   $\textbf{88.33}$  &   $\textbf{91.53}$    &   $88.04$  &   $\textbf{89.75}$\\
$\text{DeBERTa v3}_{\text{large}}$ & encoder & $434$M & \cmark & $86.75$  &   $85.83$  &   $81.95$  &   $83.85$  \\
$\text{RoBERTa}_{\text{large}}$ & encoder & $355$M & \cmark & $84.23$  &   $87.39$  &   $72.93$  &   $79.51$  \\
$\text{BERT}_{\text{large}}$ & encoder & $335$M & \cmark & $83.28$  &   $80.77$  &   $78.95$  &   $79.85$  \\
\bottomrule
\end{tabular}
\caption{Results of the classification of sentences that do not need to be judged as factually correct or incorrect in each model (binary classification). The highest value in each index is shown in bold.}
\label{tab:ex1_results}
\end{table*}

\paragraph{Sentence label annotation by AMT.}

We used Amazon Mechanical Turk (AMT) to annotate sentence labels.
The task of the crowdworker was to classify the labels of sentences (i)–(iv) by answering questions about a given sentence.
A YES/NO chart format, similar to the FaithDial creation method, was used, in which labels were classified by answering questions that can be answered with a YES/NO.
To increase data reliability, three crowdworkers were assigned per sentence, and only sentences with matching labels from two or three annotators were included in the dataset.
The following three questions were used to classify the four sentence labels.
(1) ``\textit{Is the sentence only indicating agreement/disagreement or feedback?}''
If the answer is YES, then assign label (i); if NO, then proceed to the second question.
(2) ``\textit{Is the sentence providing new information?}''
If the answer is NO, then assign label (ii); if YES, then proceed to the third question.
(3) ``\textit{Is everything in the sentence based on the speaker’s subjective opinion personal experience, thoughts, or feelings?}''
If the answer is YES, then assign label (iii); if NO, assign label (iv).
Figure~\ref{fig:AMT_Flow} shows a flowchart of the annotation process, which was also presented to the crowdworker while they were working on the task.

\subsection{Analysis of the dataset}

\paragraph{Validity of dataset annotation.}

Table~\ref{tab:annotater_label_match} shows the label match rates for the three crowdworkers assigned to each sentence during data collection.

Of the three crowdworkers assigned to each sentence, $60.0$\% of the sentences had all three labels in matching, $36.9$\% of the sentences had two labels in matching, and $3.1$\% of the sentences had all different labels and no match. 
As the percentage of no match was small, the validity of the data collection method was considered high. 
Sentences with no match were excluded from the dataset because labels could not be assigned to them.

\paragraph{Number of each labels.}

Table~\ref{tab:dataset_labels} shows the number of samples and percentage of each label in the dataset.
(iv) Objective information etc., accounted for approximately $40$\% and (iii) subjective opinions, personal experiences/thoughts/feelings, etc. accounted for approximately $40$\%.
This is possibly because when creating the base dataset WoW for FaithDial, the crowdworkers aimed to generate engaging dialogue responses by disclosing information about themselves in accordance with the statement in the instructions to ``present relevant knowledge in a fun and engaging way.''

\section{Experiment 1: Classification}

We prepared some classification models and experimentally evaluated the results of the classification (binary classification) of sentences that do not require factual correctness judgment.

\subsection{Experimental Settings}
\label{subsec:ex1_settings}
\paragraph{Dataset.}

The 1,317 collected data were divided into training and test datasets containing 1,000 and 317 responses, respectively.

\paragraph{Classification models.}
To investigate the differences in the classification accuracy based on model architecture, parameter size, and fine-tuning, experiments were conducted using $\text{GPT-3.5}$ \cite{openai2022chatgpt}, $\text{GPT-4}$ \cite{openai2023gpt4}, $\text{Llama 2}_{\text{Chat 7B}}$ \cite{touvron2023llama}, $\text{DeBERTa v3}_{\text{large}}$ \cite{he2023debertav}, $\text{RoBERTa}_{\text{large}}$ \cite{liu2019roberta}, and $\text{BERT}_{\text{large}}$~\cite{devlin-etal-2019-bert}.
Table~\ref{tab:finetuning_settings} lists our fine-tuning settings.

\paragraph{Evaluation Metrics.}

To evaluate the results of the classification of sentences that do not require factual correctness judgments (binary classification) in each model, the accuracy, precision, recall, and F1-Score were calculated.
Precision is the percentage of sentences predicted by the model as do not require factual correctness judgment and labeled as judgment not required.
Recall is the percentage of sentences labeled as factual correctness judgment not required that the model correctly predicted as sentences that do not require judgment.

\begin{table*}[ht]
    \centering
    \small
    \begin{tabular}{lcc}
        \toprule
        sentence & label & pred. \\ 
        \midrule
        \textit{My symptoms for low back pain usually improve within a few weeks if I take it easy.} & $(\mathrm{iii})$ & $1$ \\
        \midrule
        \textit{Another interesting fact about the term Blond.} & $(\mathrm{ii})$ & $1$ \\
        \midrule
        \textit{its just ashort moment of darkness before the twilight and its so inpirational} & $(\mathrm{iii})$ & $1$ \\
        \bottomrule
    \end{tabular}
    \caption{Examples of sentences that do not require a factual correctness judgment but were predicted to require one.}
    \label{tab:appendix_FN}
\end{table*}

\begin{table*}[ht]
    \centering
    \small
    \begin{tabular}{p{110mm}cc}
        \toprule
        sentence & label & pred. \\ 
        \midrule
        \textit{That means a bigger crowd.} & $(\mathrm{iv})$ & $0$ \\
        \midrule
        \textit{Reading with comprehension is very important process to learn@} & $(\mathrm{iv})$ & $0$ \\
        \midrule
        \textit{I don't know, but bamboo is the fastest growing plant in the world so I'd expect there is more than enough around to fill them up.} & $(\mathrm{iii})$ & $1$ \\
        \bottomrule
    \end{tabular}
    \caption{Examples of sentences that require a factual correctness judgment but were predicted to not require one.}
    \label{tab:appendix_FP}
\end{table*}

\subsection{Results}

Table~\ref{tab:ex1_results} shows the results of the experiment.
The highest classification accuracy was achieved with fine-tuning on the decoder model, $\text{Llama 2}_{\text{Chat 7B}}$, with an accuracy of approximately $88$ points and an F1-Score of approximately $90$ points.
For $\text{GPT-3.5}$, $\text{GPT-4}$, and $\text{Llama 2}_{\text{Chat 7B}}$ (without fine-tuning), most predictions were labels that did not require factual correctness; they had very high recall but low accuracy, precision, and F1-Score.
For the encoder models, $\text{DeBERTa v3}_{\text{large}}$ had the highest classification accuracy, whereas $\text{RoBERTa}_{\text{large}}$ and $\text{BERT}_{\text{large}}$ had almost the same accuracy.
A comparison of the decoder and encoder models with fine-tuning shows that the parameter sizes were considerably smaller for the encoder model; however, the percentage of accuracy did not differ considerably.

Tables~\ref{tab:appendix_FN} and \ref{tab:appendix_FP} show examples of sentences that could not be correctly classified by $\text{Llama 2}_{\text{Chat 7B}}$ with fine-tuning, i.e., the model with the highest classification accuracy.
Table~\ref{tab:appendix_FN} shows the examples of sentences that do not require a factual correctness judgment, but were predicted to require one, and Table~\ref{tab:appendix_FP} shows examples of sentences that required a factual correctness judgment but were predicted to not require one.

\section{Experiment 2: Relation between train data amount and accuracy}

The relation between the training data amount for fine-tuning and classification accuracy was investigated by conducting an experiment.

\subsection{Experimental Settings}

The decoder model, $\text{Llama 2}_{\text{Chat 7B}}$, and the encoder model, $\text{DeBERTa v3}_{\text{large}}$, were used in this experiment. The same settings as in Section 4.1 were used with \{$100$, $200$, $300$, $400$, $500$, $600$, $700$, $800$, $900$, $1,\!000$\} as the number of training data for fine-tuning, and the accuracy was calculated.

\begin{figure}[!ht]
    \centering
    \begin{tabular}{c}
        \begin{minipage}{\hsize}
            \centering
            \includegraphics[width=0.82\columnwidth]{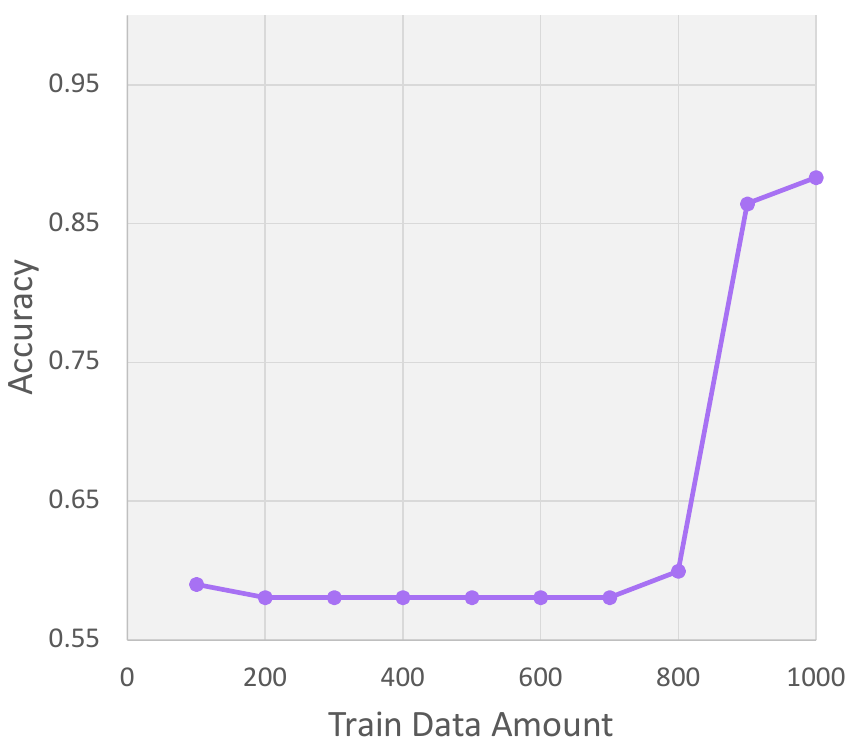}
            \hspace{20mm} (a)\hspace{1mm} $\text{Llama 2}_{\text{Chat 7B}}$
        \end{minipage}
        \\ \\
        \begin{minipage}{\hsize}
            \centering
            \includegraphics[width=0.82\columnwidth]{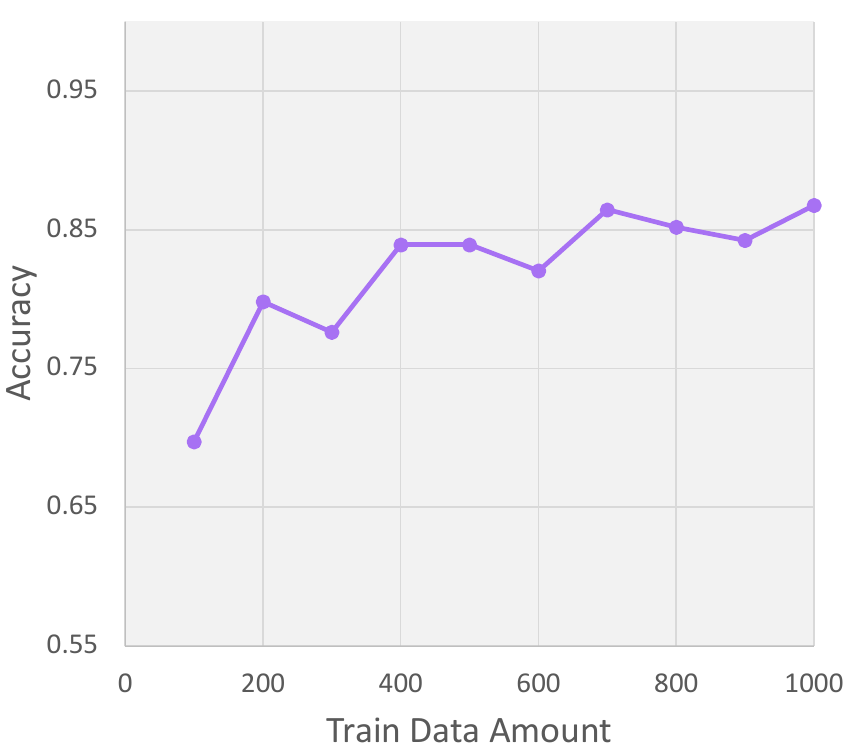}
            \hspace{20mm} (b)\hspace{1mm} $\text{DeBERTa v3}_{\text{large}}$
        \end{minipage}
    \end{tabular}
    \caption{Relationship between the amount of training data and accuracy. The accuracy of $\text{Llama 2}_{\text{Chat 7B}}$ significantly improves with over 800 training data, suggesting that more data will lead to even higher accuracy. Overall, $\text{DeBERTa v3}_{\text{large}}$ showed a steady increase in accuracy compared to $\text{Llama 2}_{\text{Chat 7B}}$.}
    \label{fig:ex2_result}
\end{figure}

\subsection{Results}

Figure~\ref{fig:ex2_result} shows the results of each model.
The accuracy rate of $\text{Llama 2}_{\text{Chat 7B}}$ increases considerably when the number of training data exceeds 800, indicating that further improvement in accuracy can be expected using additional data.
Overall, the accuracy of $\text{DeBERTa v3}_{\text{large}}$ gradually increased compared with that of $\text{Llama 2}_{\text{Chat 7B}}$.

\section{Future Directions}
\paragraph{Improving the performance of classification models.}

Herein, fine-tuning was performed on 1,000 data, which is a small amount compared to the training data size (about 18,400 responses) of the base dataset, FaithDial.
Thus, the dataset can be expanded.
As further improvement in classification accuracy can be expected by expanding the dataset, future studies will involve large-scale data collection.
It may also clarify the reason for the sudden increase in accuracy when the number of training data exceeds 800 in Figure~\ref{fig:ex2_result}(a), and whether the trend of gradual increase in accuracy in Figure~\ref{fig:ex2_result}(b) continues when training data is increased.
Moreover, because our dataset was small, the sentence labels (i), (ii), and (iii) had to be treated as a single label, “not required factual correctness judgments,” for the binary classification task.
After collecting sufficient data, we would like to investigate whether the four labels can be used for classification.

\paragraph{Application of classification models to dialogue response systems.}

If all responses that are not based on given knowledge or facts are eliminated, the attractiveness of the dialogue will be reduced.
By applying the classification models used herein, we would like to investigate whether factual and attractive dialogue responses can be generated by removing sentences related to personal feelings and opinions that do not require factual correctness judgment and then judging.

\section{Conclusion}
In this study, a task to detect sentences that do not need to be judged as factually correct or incorrect was proposed against hallucinations in a dialogue system using LLMs.
We created a dataset containing 1,317 sentences labeled with sentence types using the Amazon Mechanical Turk.
Several classification models were developed as a baseline for this task.
Results revealed that the best model could classify with an accuracy of approximately $88$\%.
In the future, we would like to collect data on a larger scale and apply the several models trained in this study to the dialogue system.

\section*{Ethics Statement}
In this study, we created datasets by human workers using a crowdsourcing platform.
In all crowdsourcing processes, identities of workers were kept anonymous and only their IDs were disclosed.
Moreover, following the terms of use of the crowdsourcing platform, an appropriate exchangeable point reward was provided for workers.

\section*{Acknowledgements}
This work was supported by JST Moonshot R\&D Grant Number JPMJMS2011-35 (fundamental research) and JSPS KAKENHI Grant Numbers JP22K17943.
We thank the Tohoku NLP Group members for their frequent discussions throughout this research and an anonymous reviewer for the insightful comments.

\newpage

\bibliography{custom}

\end{document}